\documentclass[lettersize,journal]{IEEEtran}
\usepackage{amsmath,amsfonts}
\usepackage{algorithmic}
\usepackage[T1]{fontenc}
\usepackage{mathrsfs}
\usepackage{array}
\usepackage[caption=false,font=normalsize,labelfont=sf,textfont=sf]{subfig}
\usepackage{textcomp}
\usepackage{stfloats}
\usepackage{booktabs}
\usepackage{url}
\usepackage{verbatim}
\usepackage{multicol}
\usepackage{graphicx}
\usepackage{cite}
\usepackage{color}
\usepackage{soul} 
\usepackage{color, xcolor} 
\usepackage{soul}
\sethlcolor{yellow}
\soulregister{\cite}7 
\graphicspath{{figures/}}
\def\BibTeX{{\rm B\kern-.05em{\sc i\kern-.025em b}\kern-.08em
		T\kern-.1667em\lower.7ex\hbox{E}\kern-.125emX}}
\usepackage{balance}
\begin{document}
\title{Threshold Attention Network for Semantic Segmentation of Remote Sensing Images}
\author{Wei Long, Yongjun Zhang, Zhongwei Cui, Yujie Xu, Xuexue Zhang
\thanks{
	
Wei Long, Yongjun Zhang, Yujie Xu, and Xuexue Zhang are with the State Key Laboratory of Public Big Data, College of Computer Science and Technology, Guizhou University, Guiyang 550025, Guizhou, China. (e-mail: lwsch5940@163.com, zyj6667@126.com, xuyjnobug@163.com, zhangxuexue2021@126.com)

Zhongwei Cui are with School of Mathematics and Big Data, Guizhou Education University, Guiyang 550018, China. (e-mail: zhongweicui@gznc.edu.cn).}
}


\maketitle

\begin{abstract}
	
Semantic segmentation of remote sensing images is essential for various applications, including vegetation monitoring, disaster management, and urban planning. Previous studies have demonstrated that the self-attention mechanism (SA) is an effective approach for designing segmentation networks that can capture long-range pixel dependencies. SA enables the network to model the global dependencies between the input features, resulting in improved segmentation outcomes. However, the high density of attentional feature maps used in this mechanism causes exponential increases in computational complexity. Additionally, it introduces redundant information that negatively impacts the feature representation. Inspired by traditional threshold segmentation algorithms, we propose a novel threshold attention mechanism (TAM). This mechanism significantly reduces computational effort while also better modeling the correlation between different regions of the feature map. Based on TAM, we present a threshold attention network (TANet) for semantic segmentation. TANet consists of an attentional feature enhancement module (AFEM) for global feature enhancement of shallow features and a threshold attention pyramid pooling module (TAPP) for acquiring feature information at different scales for deep features. We have conducted extensive experiments on the ISPRS Vaihingen and Potsdam datasets. The results demonstrate the validity and superiority of our proposed TANet compared to the most state-of-the-art models.
\end{abstract}


\section{Introduction}

\IEEEPARstart{R}{emote} sensing is an important source of geospatial information and plays a crucial role in numerous applications, including urban planning \cite{zheng2020parsing,marcos2018land,fu2020rotation}, vegetation monitoring \cite{running2004continuous,reed1994measuring}, military surveillance \cite{li2018aircraft}, disaster monitoring \cite{ma2019detection}, and meteorological monitoring \cite{yan2018cloud}. One of the fundamental tasks in remote sensing is semantic segmentation, which involves assigning a unique category label to each pixel in an image.

Deep learning is now widely employed in various RGB image processing tasks. FCN \cite{long2015fully} was the first fully convolutional network proposed and used in the field of semantic segmentation, implementing end-to-end pixel-level semantic segmentation. Since then, numerous networks with improvements over FCN have been proposed, including UNet, PSPNet, the Deeplab series networks, STLNet, and more. These networks typically have a two-part structure, consisting of an encoding and a decoding component. Despite the improved effectiveness of semantic segmentation achieved by these encoding-decoding network models, two important challenges still remain.

Firstly, the downsampling operation within the encoding component of a network model often leads to the loss of fine information in the original image, resulting in coarse and inaccurate predictions \cite{li2020improving}. Specifically, in regions with rich detail, such as object boundaries, the predictions tend to be particularly poor. To address this issue, a common strategy is to integrate low-level features with rich edge information into high-level features that contain more semantic information \cite{chen2018encoder,takikawa2019gated,gong2018instance,liu2020afnet,ding2020lanet}. This enhances the accuracy of the final prediction results.

Furthermore, convolutional operators in convolutional neural possess limited capability to capture long-range dependencies due to their emphasis on local features and close relationships \cite{chen20182}. The size of the receptive field provides an estimation of the amount of contextual information that can be obtained. However, the receptive field of conventional fully convolutional networks only increases linearly with the depth of the network \cite{li2020improving}.

To capture more distant dependencies in the feature map, Yu et al. \cite{yu2015multi} introduced the concept of dilated convolution, enabling the exponential growth of the receptive field without sacrificing resolution. Chen et al. \cite{chen2017deeplab} further proposed an Atrous Spatial Pyramid Pooling (ASPP) module based on multi-scale dilated convolution to extract feature information of objects at different scales. However, the use of dilated convolutions and stacked convolutional layers only provides limited contextual information, leading to limitations in modeling dependencies between distant pixels in the feature map \cite{niu2021hybrid}.

The Self-attention mechanism has been  extensively employed in tasks such as natural language processing and computer vision, owing to its potent ability to capture long-range dependencies. A prominent example of this is the Non-local network proposed by Wang et al. \cite{wang2018non}, which calculates attention weights between pixels at different locations through dot-product operations on feature maps. This integration of self-attention into the convolutional neural network enables the model to effectively capture the relationships between distant pixels.

\captionsetup[figure]{labelformat=simple, labelsep=period}
\begin{figure}[htbp]
	\centering
	\includegraphics[scale=0.5]{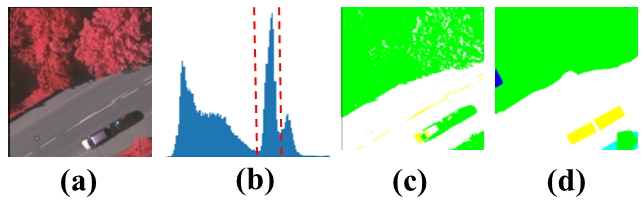}
	\caption{Traditional threshold segmentation method. (a) is the original remote sensing image. (b) is the image histogram of (a) in the red channel. (c) is the segmentation result map obtained by the traditional threshold segmentation method. (d) is the label map of (a).}
\end{figure}
	
However, this self-attention mechanism has two obvious limitations. First, generating a dense attentional feature map requires quantifying the correlation between every pixel pair, resulting in computational complexity. Second, neighboring pixels in remote sensing images are often highly correlated and their dependencies play a significant role in extracting semantic information. However, the self-attention mechanism equally considers all dependencies between pixel pairs when calculating relationships. This not only disregards local information but also introduces redundant attention weights \cite{zuo2021deformable}, resulting in a detrimental effect on feature representation \cite{niu2021hybrid}.

As illustrated in Fig. 1, the segmentation map (c) can be efficiently obtained by setting two threshold values (represented by the red dotted line) for the histogram of the red channel of the image presented in (b). The traditional threshold segmentation method possesses the advantage of aggregating all pixels with similar values across the entire image, yielding detailed edge information. However, it lacks semantic information, which leads to object misclassification, such as cars in (c) being misidentified as trees. In contrast, both neural networks and self-attention mechanisms demonstrate strong semantic information extraction capabilities.

In the process of segmenting objects in images, humans often divide the images into numerous pixel-based regions. Consequently, it is only necessary to consider the inter-block pixel relationships, rather than the relationships between individual pixels. Based on this idea, we propose the TAM. Input features are initially subjected to global quantization based on specific threshold values, generating a global threshold information vector. This vector undergoes a series of convolution and dot-product operations to compute the attention weight matrix for similar pixel aggregation regions. The final threshold attention weights are obtained by restoring the location information through another path. Compared to the self-attention mechanism, TAM not only significantly reduces computational complexity but also effectively addresses the issue of redundant dependencies between pixel pairs negatively impacting feature representation.

The primary contributions of this study are as follows:

1)
\hangafter 1
\hangindent 2.2em
\noindent
We introduce a novel TAM that focuses on the dependencies of pixel regions rather than pixel pairs. This attention mechanism provides a linear kernel attention computational complexity and effectively models the correlation between similar regions in the feature map.

2)
\hangafter 1
\hangindent 2.2em
\noindent
We develop an attentional feature enhancement module based on TAM. The AFEM modules can significantly improve the feature information of various regions where each category is located, thus obtaining an output with richer detailed features and clearer segmentation boundaries, which is beneficial for refining deeper features.

3)
\hangafter 1
\hangindent 2.2em
\noindent
We have improve the ASPP module by integrating the TAM and the enhanced ASPP module, resulting in the TAPP module. This integration enables the model to effectively capture rich global contextual information, multi-scale information, and model the relationships between similar pixel regions.

4)
\hangafter 1
\hangindent 2.2em
\noindent
We propose a novel Threshold Attention Network. TANet consists of two key components, the AFEM module and TAPP module. The AFEM module is responsible for enhancing the shallow features obtained from the image. These shallow features are then fused with deep features enhanced by the TAPP module. The resulting segmentation map is both semantically rich and finely detailed.

\section{Related Work}
\subsection{Semantic Segmentation}
Semantic segmentation, which assigns semantic labels to each pixel in an image, plays a vital role in image processing. Traditional approaches to semantic segmentation often yield suboptimal accuracy. Nevertheless, the advent of deep neural networks has facilitated considerable advancements in semantic segmentation accuracy due to their capacity for automatic extraction of more informative image features and integration of richer contextual information. Consequently, most state-of-the-art semantic segmentation algorithms presently utilize deep neural networks as their foundation.

The FCN \cite{long2015fully} was a pioneering CNN architecture that effectively performed end-to-end semantic segmentation. Subsequent to its introduction, numerous methods have been developed that build upon the innovations of FCN. For instance, U-Net \cite{chen2018encoder} introduced a symmetric encoder-decoder structure, where the encoding component extracts image features, and the decoding component recovers the edge details lost during downsampling. The ASPP module, incorporated in DeepLab \cite{chen2017deeplab}, enhances the ability of the network model to capture contextual information. STLNet \cite{zhu2021learning} leverages statistical analysis of global low-level information in feature maps to effectively extract statistical texture features at multiple scales, thus enhancing texture details. Guo et al. \cite{guo2022SegNeXt} reevaluated the characteristics necessary for successful semantic segmentation. They proposed SegNeXt, a novel convolutional attention network that utilizes inexpensive convolutional operations and achieves performance superior to transformer-based models.

Xu et al. \cite{xu2022high} proposed the High-Resolution Boundary Constraint and Context Augmentation Network (HBCNet), which utilizes boundary information, semantic information of categories, and regional feature representations to improve semantic segmentation accuracy.  CTMFNet \cite{song2022ctmfnet}, a multiscale fusion network, employs an encoder-decoder framework that integrates CNN and transformer mechanisms into its backbone network. To effectively combine local and global information in the dual backbone encoder, the authors propose a dual backbone attention fusion module (DAFM). The decoder comprises a multilayer densely connected network (MDCN), which bridges the semantic gap between scales.

\subsection{Self-attention Mechanism}
Self-attention mechanism, initially employed in the domain of NLP, has since been adopted in various other fields. Mnih et al. \cite{vaswani2017attention} combined a self-attention mechanism with a Recurrent Neural Network, allowing the network to focus on key image locations. Another notable contribution is the work of Wang et al. \cite{wang2018non}, who proposed a non-local approach using a self-attention mechanism to model interdependence between input feature map pixels.

There are two significant limitations associated with the self-attention mechanism. First, as the resolution of the input image increases, the computational burden on the network also becomes significantly large. Second, the self-attention mechanism simply computes a dot-product on the matrix that encompasses all feature information, which does not constitute a robust representation of the features.

\captionsetup[figure]{labelformat=simple, labelsep=period}
\begin{figure*}[htbp]
	\centering
	\includegraphics[scale=0.06]{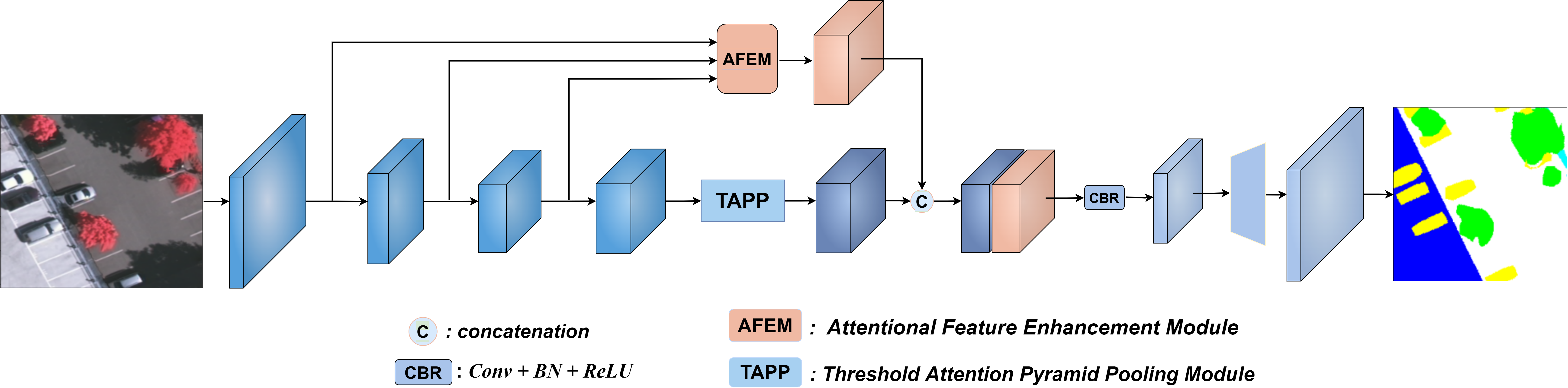}
	\caption{TANet utilizes the ResNet101 backbone network to extract features. Additionally, it employs the AFEM module to enhance the feature information obtained from the shallow network, and the TAPP module to capture rich global semantic information from the deep features. Subsequently, the two complementary feature maps are integrated to acquire a consolidated feature map. Ultimately, bilinear interpolation is leveraged to generate the ultimate predicted output map.}
\end{figure*}

Several studies have aimed to enhance the efficiency and effectiveness of self-attention mechanisms. One example is the CCNet \cite{huang2019ccnet} which employs a crossover attention mechanism to compute long-range dependencies with reduced computational cost. Another study \cite{zhang2020relation} introduced an RGA (relation-aware global attention) module to better learn attention weights by incorporating global structural information. Sun et al. \cite{sun2022spanet} proposed a SPANet with a SPAM (successive pooling attention module) that pools the value matrix to obtain features at different scales, leading to multi-scale attentional feature extraction. Guo et al. \cite{guo2022beyond} proposed a novel attention mechanism referred to as "External Attention". It incorporates two external, trainable memory modules that compute long-range dependencies between sample features to obtain attention.

We propose a novel threshold attention mechanism, which differs from prior methods in its focus on dependencies among pixel regions rather than pixel pairs. TAM aggregates features within different thresholds and applies attention to regions after aggregation. This integration of traditional threshold segmentation into self-attention reduces computational complexity and eliminates redundant noise information in the attention matrix.

\subsection{Scaling Attention Mechanism}

In addition to self-attention mechanisms, other scaled attention mechanisms have the capacity to automatically learn attention weights during the training phase, assessing the relevance of channel or spatial features. For instance, the SE module in SENet \cite{hu2018squeeze} is employed to adaptively model the interdependencies between the feature map's channels, and then the original input feature map is recalibrated based on the weights obtained for each channel. CBAM \cite{woo2018cbam} and BAM \cite{park2018bam} both model attentional weights for a given intermediate feature map in a network along both spatial and channel dimensions. However, they differ in the way they combine these weights; CBAM combines them in series, while BAM combines them in parallel. Li et al. \cite{li2021multiattention} designed a new kernel attention mechanism with linear complexity to alleviate the large computational requirements in the attention mechanism. They proposed MANet, which can combine the local feature maps extracted by the backbone network with global dependencies to adaptively weight the interdependent channel maps.

Our proposed AFEM encompasses a Channel Attention Module to dynamically learn the correlation between the feature map channels and weight coefficients. This weighting approach allows AFEM to automatically differentiate the importance of the different channels of the input features. It assigns greater weight to the more significant channels, which are crucial for achieving enhanced, detailed features.

\section{METHODOLOGY}

\subsection{Overview}
The TANet is introduced with its overall structure in Fig. 2. For feature extraction during the encoding phase, the backbone network is ResNet101 with dilated convolution. The shallow network output provides abundant details but lacks semantic information, whereas the deep network output offers rich semantic information but lacks details. The three shallow feature maps obtained from the encoding phase are concatenated and fed into the AFEM to get a feature map with both enhanced detailed information. The deep features from the backbone network are input into the TAPP to obtain a feature map with rich semantic and contextual information. These complementary feature maps are then concatenated and fused to get a unified feature map. Finally, bilinear interpolation is utilized to obtain the ultimate prediction output map, which possesses the same dimensions as the input image.

The TANet integrates global contextual information and feature information at various scales to produce high-level features enriched with semantic information. Moreover, TANet also improves low-level features, which are replete with detail but devoid of semantic information. The progressive fusion of these two types of features results in a more precise and detailed segmentation prediction map.

\subsection{Threshold Attention Mechanism}
The threshold segmentation method is a widely used algorithm in conventional image segmentation. This method is based on the principle that the pixel values of different objects in an image are significantly distinct. To obtain the required pixel thresholds, a calculation can be performed or the pixel statistical histogram of the image can be processed. Subsequently, the pixels in the image are classified based on these thresholds, resulting in the segmentation of the different objects present in the image.

\captionsetup[figure]{labelformat=simple, labelsep=period}
\begin{figure*}[htbp]
	\centering
	\includegraphics[scale=0.11]{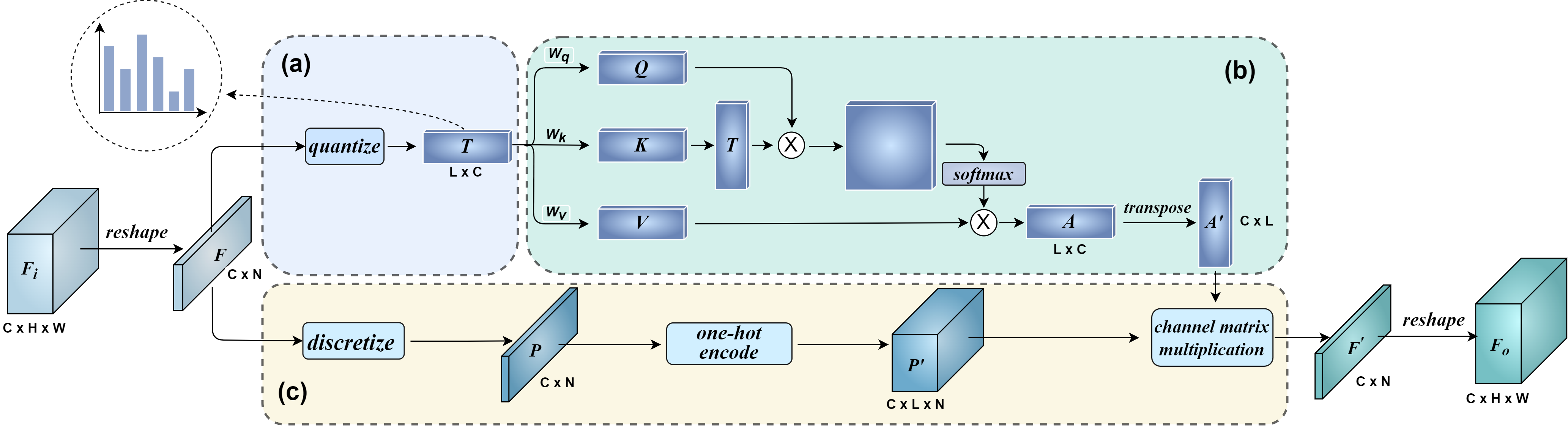}
	\caption{TAM consists of three parts: (a) for thresholding the input features, (b) for calculating the attention weight matrix, and (c) for recovering location information for features. TAM is an attention mechanism that exhibits linear computational complexity and effectively models the correlation between similar regions in the feature map.}
\end{figure*}

Inspired by this traditional approach, we present the Threshold Attention Mechanism, which learns attention weights for different pixel regions in a feature map. To achieve this, we quantize each channel of the input feature map with a global threshold information matrix, resulting in a threshold feature matrix. This matrix undergoes convolution and dot-product calculations to obtain an attention weight matrix. The input feature map is also discretized into feature classes to form a position matrix. By multiplying the attention weight matrix and position matrix, we get an output matrix that assigns attention weights to different pixel regions. The figure in Fig. 3 depicts a graphical representation of TAM.

\textit{\textbf{a)} Thresholding the input features:}

Define the input features as $\textbf{\textit{F}}_{i}\in \mathbb{R}^{C\times H\times W}$. and then reshape the features into $\textbf{\textit{F}}\in \mathbb{R}^{C\times N}$, where $N=H\times W$. Quantize each channel in $\textbf{\textit{F}}$ separately using a threshold that is based on the feature data distribution in the different channels. This quantization enables the grouping of pixels with similar characteristics in the original feature into disparate threshold clusters.

\begin{equation}
	\textbf{\textit{T}}_{c,l}=\frac{max(\textbf{\textit{F}}_{c})-min(\textbf{\textit{F}}_{c})}{2\textbf{\textit{L}}}\times (2l-1)+min(\textbf{\textit{F}}_{c})
\end{equation}

where $c\in [1,C]$ and $l\in [1,L]$. $\textbf{\textit{F}}_{c}$ denotes the feature data of the $c$-th channel in matrix $\textbf{\textit{F}}$. $\textbf{\textit{L}}$ represents the number of feature levels to be quantized, which implies that the data of every channel will be divided into $\textbf{\textit{L}}$ intervals of equal size based on a certain threshold. $\textbf{\textit{T}}_{c,l}$ denotes the quantization result when the feature of the $c$-th channel in the input matrix $F$ is within the $l$-th threshold, and the feature matrix $\textbf{\textit{T}}\in \mathbb{R}^{L\times C}$ is obtained after this quantization operation.

\textit{\textbf{b)} Calculating the attention weight matrix:}

Similar to the dot-product attention, we use three different projection matrices, $\textbf{\textit{W}}_{q}\in\mathbb{R}^{C\times C}$, $\textbf{\textit{W}}_{k}\in\mathbb{R}^{C\times C}$, and $\textbf{\textit{W}}_{v}\in\mathbb{R}^{C\times C}$, to generate the corresponding query matrix $\textbf{\textit{Q}}$, key matrix $\textbf{\textit{K}}$, and value matrix $\textbf{\textit{V}}$.

\begin{equation}
	\textbf{\textit{Q}}=\textbf{\textit{T}}\textbf{\textit{W}}_{q}\in \mathbb{R}^{L\times C}
\end{equation}
\begin{equation}
	\textbf{\textit{K}}=\textbf{\textit{T}}\textbf{\textit{W}}_{k}\in \mathbb{R}^{L\times C}
\end{equation}
\begin{equation}
	\textbf{\textit{V}}=\textbf{\textit{T}}\textbf{\textit{W}}_{v}\in \mathbb{R}^{L\times C}
\end{equation}
\begin{equation}
	\rho (\textbf{\textit{Q}}\textbf{\textit{K}}^{T})=softmax_{row}(\textbf{\textit{Q}}\textbf{\textit{K}}^{T})
\end{equation}

We employ a normalization function $\rho$ to measure the similarity between the $i$-th query feature  $\textbf{\textit{q}}_{i}^{T}\in \mathbb{R}^{C}$ and the $j$-th key feature $\textbf{\textit{k}}_{j}\in \mathbb{R}^{C}$, i.e., $\rho (\textbf{\textit{q}}_{i}^{T}\cdot \textbf{\textit{k}}_{j})\in \mathbb{R}^{1}$. This matrix $\textbf{\textit{Q}}\textbf{\textit{K}}^{T}$ models the dependencies between different threshold features (features in different pixel regions) for different channels in the threshold feature matrix $\textbf{\textit{T}}$. Obtain the attention matrix by first normalizing the attention weight values in the relationship matrix $\textbf{\textit{Q}}\textbf{\textit{K}}^{T}$ via the Softmax function (denoted as $softmax_{row}$). Then, generate the attention matrix by re-weighting $V$ with the normalized attention weight values.

\begin{equation}
	\textbf{\textit{A}}=\rho (\textbf{\textit{Q}}\textbf{\textit{K}}^{T})\textbf{\textit{V}}
\end{equation}

\textit{\textbf{c)} Recovering location information for features:}

Matrix $\textbf{\textit{A}}$ holds global dependency information from input matrix $\textbf{\textit{F}}$ but lacks corresponding location information for each pixel feature due to prior quantization. As shown in Fig. 3, in order to retrieve this information, a "discretize" operation is performed on matrix F to obtain matrix $\textbf{\textit{P}}$, which records the quantization levels for all pixel features. This allows for the reconstruction of the original pixel location information in input feature $\textbf{\textit{F}}$.

\begin{equation}
	{\textbf{\textit{P}}}_{c,n}=\left \lfloor \frac{\textit{\textbf{F}}_{c,n}-min({\textbf{\textit{F}}}_{c})}{max({\textbf{\textit{F}}}_{c})-min({\textbf{\textit{F}}}_{c})} \times 2\textbf{\textit{L}} \right \rfloor
\end{equation}

where $n\in [1, N]$. $\textbf{\textit{F}}_{c,n}$ is the $n$-th feature of the $c$-th channel of the input feature matrix $\textbf{\textit{F}}$. $\textbf{\textit{P}}_{c,n}$ is the integer feature value obtained by thresholding $\textbf{\textit{F}}_{c,n}$ according to $\textbf{\textit{L}}$. 

From this, we can obtain a matrix $\textbf{\textit{P}}\in R^{C\times N}$ that records the location information of the corresponding quantization level of each pixel feature of the matrix $\textbf{\textit{F}}$. The matrix $\textbf{\textit{P}}$ is then one-hot encoded to obtain the matrix $\textbf{\textit{P}}{}'$, and the $\textbf{\textit{A}}$ matrix is transposed to obtain $\textbf{\textit{A}}{}'$.

\begin{equation}
	\textbf{\textit{F}}{}'_{c}=\textbf{\textit{A}}{}'_{c}\cdot \textbf{\textit{P}}{}'_{c}
\end{equation}

where $\textbf{\textit{A}}{}'_{c}\in \mathbb{R}^{1\times L}$ and $\textbf{\textit{P}}{}'_{c}\in \mathbb{R}^{L\times N}$ are the vectors of the $c$-th channel in the matrices $\textbf{\textit{A}}{}'_{c}$ and $\textbf{\textit{P}}{}'_{c}$ respectively. The threshold attention weights are reassigned to the features by multiplying the $\textbf{\textit{A}}{}'_{c}$ and $\textbf{\textit{P}}{}'_{c}$ matrices. The resulting matrix $\textbf{\textit{F}}{}'_{c}\in \mathbb{R}^{C\times N}$ undergoes a reshape operation to obtain the final output feature matrix $\textbf{\textit{F}}_{o}\in \mathbb{R}^{C\times H\times W}$ of the TAM module. It is worth noting that the input and output features of the TAM module have the same shape size ($C\times H\times W$).

The TAM module models the correlation between sets of features that lie within distinct thresholds of the input feature map, thereby capturing the dependencies between blocks of homogeneous pixel regions. This leads to a dynamic assignment of attention to various sets of pixels, thereby enhancing the features of the input matrix.

\subsection{Attentional Feature Enhancement Module}
Our proposed Attentional Feature Enhancement Module comprises three branches. As depicted in Fig. 4, one branch conducts channel attention acquisition through global averaging pooling of the input feature map and two fully connected layers. Another branch, the TAM, computes the cosine similarity of the input feature map with a globally averaged feature vector. It then models correlations among similar regions to enhance the feature map with attentional features. The channel attention weights learned by the first branch are applied to the feature map obtained from TAM. The third branch, based on residual connectivity, adds the original feature maps to those from TAM and channel attention weight assignment, enabling the network to automatically learn feature assignment and facilitate gradient back-propagation. The AFEM produces a feature map of the same size as the input, rendering it easy to integrate into the network.

\captionsetup[figure]{labelformat=simple, labelsep=period}
\begin{figure}[htbp]
	\centering
	\includegraphics[scale=0.08]{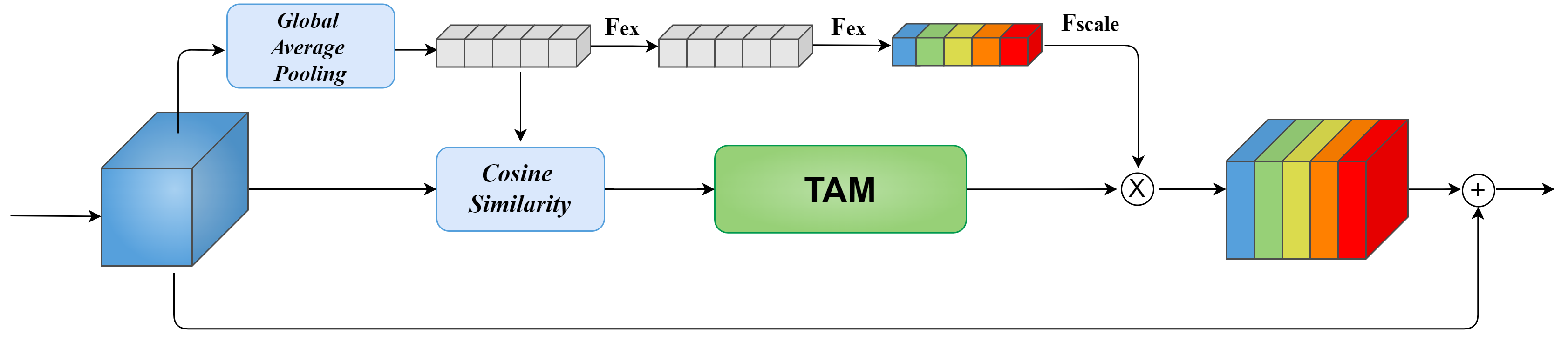}
	\caption{The structure of the AFEM is composed of three branches: the first one acquires channel attention, the second one enhances threshold attentional features, and the third one provides residual connectivity.}
\end{figure}

\subsection{Threshold Attention Pyramid Pooling}
The ASPP module employs four parallel dilation convolutions to construct features with varying perceptual fields, which enhances information extraction of objects at different scales in the image. However, this approach may lead to a loss of detail information and insufficient global feature relevance information. To address these limitations, we propose the Threshold Attention Pyramid Pooling method. TAPP improves ASPP by increasing the convolution kernel size for a larger perceptual field. Additionally, it adds a threshold-attention branch to model correlations between similar pixel regions, resulting in rich global contextual information with low computational cost.

Fig. 5 shows the threshold attention space pooling module divided into three branches: expansion convolution (with varying expansion rates), global average pooling, and threshold attention. The expansion convolution branch extracts multi-scale features in parallel using 3 dilation convolutions of sizes 4, 6, and 8 (K is both kernel size and expansion rate). To reduce computation, we use depth-wise convolution with 1xK and Kx1 dilation convolutions. The threshold attention branch computes input feature similarity (Cos) with GAP-computed features, and inputs the results into TAM to calculate attention weights for different pixel regions. In addition, a convolution kernel size of 1 is added to both the Global Average Pooling (GAP) and TAM branches to reduce feature dimensionality and improve feature representation.

\captionsetup[figure]{labelformat=simple, labelsep=period}
\begin{figure}[htbp]
	\centering
	\includegraphics[scale=0.11]{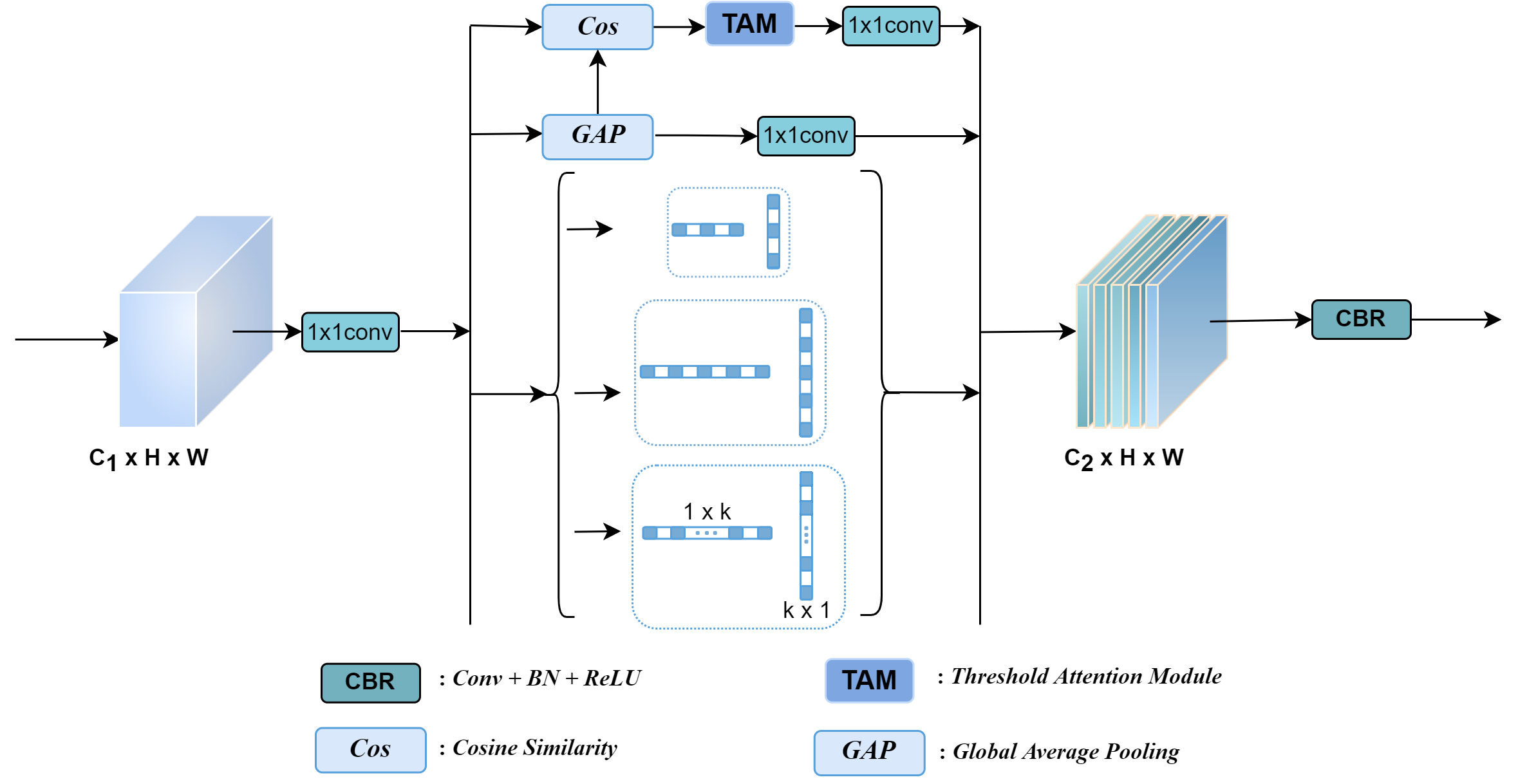}
	\caption{Structure of the TAPP, where CBR is the convolution layer + BN layer + ReLU layer, Cos is the calculation of cosine similarity, and GAP is the calculation of global average pooling.}
\end{figure}

\subsection{Loss Functions}
For supervised training, we selected cross-entropy loss as the main predictive loss. Its formula is as follows:

\begin{equation}
	\mathcal{L}_{CE}=-\frac{1}{N}\sum_{n=1}^{N}\sum_{k=1}^{K}y_{k}^{(n)}log\hat{y}_{k}^{(n)}
\end{equation}

Where $n\in [1,2,..., N]$, $N$ is the number of samples, and $K$ is the number of categories. $\hat{y}_{k}^{(n)}$ is the one-hot vector of the network's output after softmax, and $y_{k}^{(n)}$ is the true label value corresponding to this sample.

Supervised training using only the difference between final layer output and true label maps slows convergence and yields limited results. To resolve this, we added an auxiliary loss in the third block of the backbone network, using cross-entropy loss as in the final layer.

\begin{equation}
	\mathcal{L}=\mathcal{L}_{CE}+\lambda \mathcal{L}_{aux}
\end{equation}

Where $\mathcal{L}_{CE}$ is the prediction loss, and $\mathcal{L}_{aux}$ is the auxiliary loss. Hyperparameter $\lambda$ balances the weights between primary and auxiliary losses. Main loss uses online hard examples mining (OHEM) to focus network learning on difficult-to-classify pixels with prediction vector probability less than $\theta $. Hard-to-classify pixels have individual losses calculated and back-propagated for network optimization. Instead of conducting an exhaustive search for optimal parameter values, we determined values that produced stable segmentation effects through limited experiments on the Vaihingen dataset. Our selected parameter values were $\lambda$=0.5, $\theta$=0.65, and S=10,000.

\section{Datasets and Experimental Settings}

\subsection{Datasets}
We evaluated the efficacy of our proposed Threshold Attention Network through experiments on two well-known open datasets: the ISPRS Vaihingen dataset and the ISPRS Potsdam dataset. Both datasets include six classes of remote sensing image labels: ground, building, low vegetation, tree, vehicle, and background (clutter). The ISPRS dataset provides two types of semantic labels for testing, one with eroded boundaries and one without. In our experiments, we used the semantic labels with eroded boundaries

\textit{1) Vaihingen Dataset:}
The ISPRS Vaihingen dataset contains 33 very high-resolution orthophoto maps. The average size of the images in the dataset is 2494$\times$2064 pixels. The orthophoto images have three channels: the infrared channel, the red channel, and the green channel, each containing a wealth of spectral information. In addition, there are two sets of ancillary data in the dataset: the Digital Surface Model (DSM) and the Normalised Digital Surface Model data (NDSM). This dataset is formally divided into 16 training regions and 17 test regions. For the partitioning of the dataset, our setup is the same as \cite{fu2019dual,chen2017rethinking}, with the 15 images labeled as follows: 2, 4, 6, 8, 10, 12, 14, 16, 20, 22, 24, 27, 29, 31, 33, 35, and 38 selected for training. Thirty labeled images are used for validation, and the remaining 17 images are used as the test set.

In our experiments, we did not utilize DSM or NDSM data. To train the network model, we preprocessed the remote sensing images by cropping them to 512 $\times$ 512, and data augmentation techniques were applied including random rotation ($90^{\circ}$, $180^{\circ}$, $270^{\circ}$), random resizing (0.5-2), the addition of random Gaussian noise, and random horizontal and vertical flipping.

\textit{2) Potsdam Dataset:}
The Potsdam dataset comprises 38 fine-resolution images, all 6000 x 6000 pixels in size. The dataset includes NIR, red, green, and blue channels, as well as DSM and normalized DSM (NDSM) data. We divided it in the same way as \cite{chen2017rethinking}, using the 14 images labeled: 2\_13, 2\_14, 3\_13, 3\_14, 4\_13, 4\_14, 4\_15, 5\_13, 5\_14, 5\_15, 6\_13, 6\_14, 6\_15, and 7\_13 for testing, ID: 2\_10 for validation, and except for image 7\_10 with incorrect annotations 22 images were utilized as the training set. As with the Vaihingen dataset, we did not use the DSM and NDSM data. We used image cutting and data enhancement in the same way as on the Vaihingen dataset.

\subsection{Evaluation Metrics}
The performance of TANet was evaluated using three metrics: overall accuracy (OA), mean intersection over union (mIoU), and F1 score. Based on the cumulative confusion matrix, these evaluation metrics are calculated as:

\begin{equation}
	OA=\frac{\sum _{k=1}^{N}TP_{k}}{\sum _{k=1}^{K}TP_{k}+FP_{k}+TN_{k}+FN_{k}}
\end{equation}

\begin{equation}
	mIoU=\frac{1}{N}\sum_{k=1}^{N}\frac{TP_{k}}{TP_{k}+FP_{k}+FN_{k}}
\end{equation}

\begin{equation}
	precision_{k}=\frac{TP_{k}}{TP_{k}+FP_{k}}
\end{equation}

\begin{equation}
	recall_{k}=\frac{TP_{k}}{TP_{k}+FN_{k}}
\end{equation}

\begin{equation}
	F1_{k}=2\times \frac{precision_{k}\times recall_{k}}{precision_{k}+recall_{k}}
\end{equation}

Where $TP_{k}$, $TN_{k}$, $FN_{k}$ and $FP_{k}$ denote true positives, false positives, true negatives and false negatives respectively for a particular object indexed as category $k$.

\subsection{Implementation Details}

For all comparisons, we employ ResNet-101 pre-trained on the ImageNet dataset as the backbone network. The final two downsampling operations are replaced with dilated convolutional layers with expansion rates of 2 and 4 \cite{zhao2017pyramid}, resulting in an output stride of 8. The AdamW optimizer, which includes weight decay, is used. During training, a 'poly' strategy is applied to set the learning rate, calculated as the initial learning rate multiplied by $(1-\frac{max\_iter}{iter})^{0.9}$, with an initial value of 0.0005. Experiments were conducted on an NVIDIA Tesla V100 GPU with 32 GB memory. The threshold number L of the threshold attention module was optimized for different datasets.

\section{Experimental Results and Analysis}
\subsection{Parameter Study for the TANet}
The proposed threshold attention module has a crucial parameter, $L$, referred to as the threshold number. This parameter determines the level of granularity in the attention applied to the input features. We experimentally studied the effect of $L$ on the segmentation performance of the network. We studied the effect of threshold values $L_1$ and $L_2$ in AFEM and TAPP modules on the Vaihingen and Potsdam datasets, respectively. We first set $L_2$ in TAPP to 200 and sought the optimal value of $L_1$ in AFEM. Next, we found the optimal value of $L_2$. Additionally, we evaluated the impact of incorporating dilated convolutions of varying scales into the TAPP module on the model's segmentation performance via an experiment.

\begin{table}[htbp]
	\caption{\textsc{Results of Ablation Experiments on the Vaihingen Dataset for the $L_1$ Parameter in the AFEM Module}}
	\scalebox{1.05}{
		\begin{tabular}{c|cccccc}
			\toprule
			$L_1$ 		& 50 	& 100 	& 150 	         & 200   & 250 	& 300 \\
			\midrule
			Mean F1(\%) & 90.54 & 90.66 & \textbf{90.78} & 90.64 & 90.50 & 90.51 \\
			OA(\%)      & 90.97 & 90.94 & \textbf{91.13} & 91.03 & 90.77 & 90.90 \\
			mIoU(\%)    & 82.94 & 83.17 & \textbf{83.35} & 83.15 & 82.91 & 82.90 \\
			\bottomrule
	\end{tabular}}
\end{table}

\begin{table}[htbp]
	\caption{\textsc{Results of Ablation Experiments on the Vaihingen Dataset for the $L_2$ Parameter in the TAPP Module}}
	\scalebox{1.05}{
		\begin{tabular}{c|cccccc}
			\toprule
			$L_2$ 		& 50 	& 100 	& 150 	& 200    & 250   & 300 \\
			\midrule
			Mean F1(\%) & 90.37 & 90.51 & 90.69 & \textbf{90.78} & 90.64 & 90.45 \\
			OA(\%)      & 90.77 & 91.13 & 90.97 & \textbf{91.13} & 91.11 & 91.07 \\
			mIoU(\%)    & 82.67 & 82.91 & 83.22 & \textbf{83.35} & 83.13 & 82.81 \\
			\bottomrule
	\end{tabular}}
\end{table}

\begin{table}[hp]
	\caption{\textsc{Results of ablation experiments on the Vaihingen dataset for dilated convolution in TAPP}}
	\scalebox{1.02}{
		\begin{tabular}{c|ccc|c|c|c}
			\toprule
			Method & 4x4       & 6x6        & 8x8        & Mean F1(\%)    & OA(\%)		  & mIoU(\%) \\
			\midrule
			TANet & 		   &			&		   	 & 90.57          & 90.90		  & 82.98 \\
			TANet & \checkmark &			&			 & 90.62          & 90.93 		  & 83.11 \\
			TANet & \checkmark & \checkmark &			 & 90.71          & 91.10  		  & 83.24 \\
			TANet & \checkmark & \checkmark & \checkmark & \textbf{90.78} & \textbf{91.13} & \textbf{83.35} \\
			\bottomrule
	\end{tabular}}
\end{table}

\textit{1) Experiments on the Vaihingen dataset}

The results of the experiment on the Vaihingen dataset are presented in Tables I and II. It is evident that TANet obtains the optimal semantic segmentation performance when $L_1$ is set to 150 and $L_2$ is set to 200. Furthermore, it is observed that the model demonstrates a greater sensitivity to the parameter $L_2$ as compared to $L_1$. Table III displays the outcomes of the ablation experiments on the Vaihingen dataset for the dilated convolution in TAPP. The results reveal that the use of all three scales of the dilated convolution enhances segmentation performance, with the optimal results obtained when all three scales are utilized together.

\textit{2) Experiments on the Potsdam dataset}

The results for the Potsdam dataset are presented in Tables IV and V. The optimal semantic segmentation is achieved when both $L_1$ and $L_2$ are set to 200. No significant difference in the sensitivity to parameters $L_1$ and $L_2$ was observed. Similar to the results obtained on the Vaihingen dataset, all three scales of the dilated convolution in Table VI help to improve the final segmentation performance of TANet.

\subsection{Ablation Study}

\textit{1) Comparison with context aggregation modules and attention modules}

Table VII compares our proposed model with classical context extraction modules and four newer attention mechanisms in terms of segmentation effectiveness. The results indicate that the AFEM and TAPP modules achieve better segmentation accuracy than other context extraction modules and attention mechanisms. Our proposed thresholded attention results in a more effective extraction of attentional information by modeling the feature correlation among different pixel regions. The experimental results proved the effectiveness of the method.

\begin{table}[htbp]
	\caption{\textsc{Results of Ablation Experiments on the Potsdam Dataset for the $L_1$ Parameter in the AFEM Module}}
	\scalebox{1.08}{
		\begin{tabular}{c|cccccc}
			\toprule
			$L_1$       & 50    & 100   & 150   & 200            & 250   & 300 \\
			\midrule
			Mean F1(\%) & 93.16 & 93.19 & 93.30 & \textbf{93.35} & 93.29 & 93.13 \\
			OA(\%)      & 91.79 & 91.85 & 91.94 & \textbf{92.10} & 91.93 & 91.63 \\
			mIoU(\%)    & 87.43 & 87.48 & 87.68 & \textbf{87.75} & 87.64 & 87.41 \\
			\bottomrule
	\end{tabular}}
\end{table}

\begin{table}[htbp]
	\caption{\textsc{Results of Ablation Experiments on the Potsdam Dataset for the $L_2$ Parameter in the TAPP Module}}
	\scalebox{1.08}{
		\begin{tabular}{c|cccccc}
			\toprule
			$L_2$       & 50    & 100   & 150   &	  200        & 250 & 300 \\
			\midrule
			Mean F1(\%) & 93.13 & 93.17 & 93.16 & \textbf{93.35} & 93.15 & 93.13 \\
			OA(\%)      & 91.79 & 91.87 & 91.79 & \textbf{92.10} & 91.76 & 91.80 \\
			mIoU(\%)    & 87.39 & 87.45 & 87.45 & \textbf{87.75} & 87.41 & 87.38 \\
			\bottomrule
	\end{tabular}}
\end{table}

\begin{table}[htbp]
	\caption{\textsc{Results of ablation experiments on the Potsdam dataset for dilated convolution in TAPP}}
	\scalebox{1.02}{
		\begin{tabular}{c|ccc|c|c|c}
			\toprule
			Method & 4x4       & 6x6        & 8x8        & Mean F1(\%)    & OA(\%)		  & mIoU(\%) \\
			\midrule
			TANet & 		   &			&		   	 & 93.04          & 91.54 		  & 87.22 \\
			TANet & \checkmark &			&			 & 93.10          & 91.70 		  & 87.34 \\
			TANet & \checkmark & \checkmark &			 & 93.28          & 91.99  		  & 87.65 \\
			TANet & \checkmark & \checkmark & \checkmark & \textbf{93.35} & \textbf{92.10} & \textbf{87.75} \\
			\bottomrule
	\end{tabular}}
\end{table}

The improvement effect of the TAPP module and AFEM module on the network model is comparable. Both TAPP and AFEM modules effectively enhance the semantic segmentation performance of the model. The combination of these two modules and the baseline network results in TANet, which achieves the best segmentation results.

\begin{table}[htbp]
	\caption{\textsc{Results of the Ablation Experiments on the Vaihingen Dataset}}
	\scalebox{0.97}{
		\begin{tabular}{l|c|c|c}
			\toprule
			Method &  Mean F1(\%) &  OA(\%) & mIoU(\%) \\
			\midrule
			ResNet-101 Baseline   & 89.84 & 90.38 & 81.82 \\
			\midrule
			ResNet-101+SE \cite{hu2018squeeze}         & 90.11 & 90.80 & 82.26 \\
			ResNet-101+SA \cite{wang2018non}           & 90.16 & 90.70 & 82.30 \\
			ResNet-101+ASPP \cite{chen2017rethinking}  & 90.23 & 90.88 & 82.46 \\
			ResNet-101+DAB \cite{fu2019dual}           & 90.33 & 90.97 & 82.60 \\
			ResNet-101+PPM \cite{zhao2017pyramid}      & 90.34 & 90.80 & 82.62 \\
			ResNet-101+EA \cite{guo2022beyond}         & 90.41 & 90.77 & 82.75 \\
			ResNet-101+CAM\&KAM \cite{li2021multiattention} & 90.41 & 90.96 & 82.71 \\
			ResNet-101+BCM\&CEM \cite{xu2022high}      & 90.44 & 90.82 & 82.76 \\
			ResNet-101+CAA\&RSA \cite{niu2021hybrid}        &   -   & 90.98 & 82.87 \\
			\midrule
			ResNet-101+TAM        & 90.58 & 90.90 & 83.03 \\
			ResNet-101+AFEM       & 90.63 & 91.09 & 83.14 \\
			ResNet-101+TAPP       & 90.65 & 90.99 & 83.15 \\
			ResNet-101+TAPP\&AFEM (ours)  & \textbf{90.78} & \textbf{91.13} & \textbf{83.35} \\
			\bottomrule
	\end{tabular}}
\end{table}

\begin{table}[htbp]
	\caption{\textsc{Efficiency Comparison With Context Aggregation Modules And Attention Modules When Processing Input Feature Map Of Size ($1 \times 2048 \times 128 \times 128$) During The Inference Stage}}
	\scalebox{1.1}{
		\begin{tabular}{l|c|c|c}
			\toprule
			Method     & GFLOPs & Params(M)  & Memory(MB) \\
			\midrule
			LKPP \cite{zheng2020parsing}       & 884     & 54.5 & 818 \\
			PPM \cite{zhao2017pyramid}       & 619     & 22.0 & 792 \\
			ASPP \cite{chen2017rethinking}      & 503     & 15.1 & 284 \\
			CCA \cite{deng2021ccanet}       & 804     & 10.6 & 427 \\
			SA \cite{wang2018non}         & 619     & 10.5 & 2168 \\
			OCR \cite{yuan2020object}       & 354     & 10.5 & 202 \\
			CAA\&RSA \cite{niu2021hybrid}       & 292     & 13.1  & 393 \\
			PAM\&AEM \cite{ding2020lanet}  & 158     & 10.4 & 489 \\
			CAM\&KAM \cite{li2021multiattention}  & 86      & 5.3  & \textbf{160} \\
			\midrule
			AFEM\&TAPP (ours) & \textbf{49} & \textbf{4.5}  & 262 \\
			
			\bottomrule
	\end{tabular}}
\end{table}

\begin{table}[htbp]
	\caption{\textsc{Results Of Inference Time Comparison Between Tanet And Other Models}}
	\scalebox{1.1}{
		\begin{tabular}{l|c}
			\toprule
			Method     						& Average inference time per image (seconds) \\
			\midrule
			FCN 		\cite{long2015fully}       		& $0.071\pm0.001$     	\\
			SA 			\cite{wang2018non}      		& $0.076\pm0.001$      	\\
			PSPNet 		\cite{zhao2017pyramid}       	& $0.084\pm0.001$  		\\
			EANet 		\cite{guo2022beyond}       		& $0.086\pm0.001$    	\\
			SENet 		\cite{hu2018squeeze}       		& $0.088\pm0.001$  		\\
			DeeplabV3+ 	\cite{chen2017rethinking}       & $0.088\pm0.001$   	\\
			DABNet 		\cite{fu2019dual}      		 	& $0.097\pm0.001$  		\\
			\midrule
			TANet (ours) 								& $0.091\pm0.001$ 		\\
			\bottomrule
	\end{tabular}}
\end{table}

\begin{table}[htbp]
	\caption{\textsc{Results of Ablation Experiments with Different Improvements on the Vahingen Dataset}}
	\scalebox{0.9}{
		\begin{tabular}{c|ccc|c|c|c}
			\toprule
			Method & OHEM      & Aux Loss   & TTA       & Mean F1(\%) & OA(\%) & mIoU(\%) \\
			\midrule
			TANet & 		   &			&			& 90.78       & 91.13  & 83.35 \\
			TANet & \checkmark &			&			& 90.85       & 91.23  & 83.46 \\
			TANet & \checkmark & \checkmark &			& 91.16       & 91.50  & 83.99 \\
			TANet & \checkmark & \checkmark & \checkmark & \textbf{91.45} & \textbf{91.93} & \textbf{84.45} \\
			\bottomrule
	\end{tabular}}
\end{table}

\begin{table}[htbp]
	\caption{\textsc{Results of Ablation Experiments with Different Improvements on the Potsdam Dataset}}
	\scalebox{0.9}{
		\begin{tabular}{c|ccc|c|c|c}
			\toprule
			Method & OHEM      & Aux Loss   & TTA       & Mean F1(\%) & OA(\%) & mIoU(\%) \\
			\midrule
			TANet & 		   &			&			& 93.35       & 92.10  & 87.75 \\
			TANet & \checkmark &			&			& 93.40       & 92.27  & 87.85 \\
			TANet & \checkmark & \checkmark &			& 93.52       & 92.32  & 88.06 \\
			TANet & \checkmark & \checkmark & \checkmark & \textbf{93.71} & \textbf{92.95} & \textbf{88.43} \\
			\bottomrule
	\end{tabular}}
\end{table}

\begin{table*}[htbp]
	\caption{\centering \textsc{Quantitative Comparisons with State of the Arts on the Vaihingen Test Set}}
	\scalebox{1.05}{
		\begin{tabular}{l|c|c|cccc|c|c|c}
			\toprule
			& Backbone & Imp.surf &Building & Low veg & Tree  & Car & Mean F1(\%) & OA(\%) & mIoU(\%) \\
			\midrule
			V-FuseNet \cite{audebert2018beyond}		& 		-	    & 92.00 & 94.40 & 84.50 & 89.90 & 86.30 & 89.42 & 90.00 & -     \\
			DLR\_9 \cite{marmanis2018classification}& 		-	    & 92.40 & 95.20 & 83.90 & 89.90 & 81.20 & 88.52 & 90.30 & -     \\
			TreeUNet \cite{yue2019treeunet}			& 		-	    & 92.50 & 94.90 & 83.60 & 89.60 & 85.90 & 89.30 & 90.40 & -     \\
			DANet \cite{fu2019dual}					& 	ResNet-101  & 91.63 & 95.02 & 83.25 & 88.87 & 87.16 & 89.19 & 90.44 & 81.32 \\
			DeepLabV3+ \cite{chen2017rethinking}	&  	ResNet-101	& 92.38 & 95.17 & 84.29 & 89.52 & 86.47 & 89.57 & 90.56 & 81.47 \\
			ABCNet \cite{li2021abcnet}				& 	ResNet-18   & 92.70 & 95.20 & 84.50 & 89.70 & 85.30 & 89.50 & 90.70 & 81.30 \\
			PSPNet \cite{zhao2017pyramid}			&   ResNet-101	& 92.79 & 95.46 & 84.51 & 89.94 & 88.61 & 90.26 & 90.85 & 82.58 \\
			ACFNet \cite{zhang2019acfnet}			& 	ResNet-101  & 92.93 & 95.27 & 84.46 & 90.05 & 88.64 & 90.27 & 90.90 & 82.68 \\
			MANet \cite{li2021multiattention}		& 	ResNet-101  & 93.02 & 95.47 & 84.64 & 89.98 & 88.95 & 90.41 & 90.96 & 82.71 \\	
			CASIA2 \cite{liu2018semantic}			& 	ResNet-101  & 93.29 & 96.00 & 84.70 & 89.90 & 86.70 & 90.10 & 91.10 & -     \\
			CCANet \cite{deng2021ccanet}			& 	ResNet-101  & 93.29 & 95.53 & 85.06 & 90.34 & 88.70 & 90.58 & 91.11 & 82.76 \\
			HMANet \cite{niu2021hybrid}				& 	ResNet-101  & 93.50 & 95.86 & 85.41 & 90.40 & 89.63 & 90.96 & 91.44 & 83.49 \\
			MFNet \cite{su2022semantic}				& 	ResNet-50   & 93.43 & 96.35 & 85.85 & 90.50 & 88.31 & 90.88 & 91.67 & 83.50 \\
			
			CTMFNet \cite{song2022ctmfnet}		& 	HRNet\&transformer  & 93.79 & 96.12 & 85.02 & 90.47 & \textbf{91.47} & 91.37 & 91.60 & 84.34 \\
			DC-Swin \cite{wang2022novel}		& 	Swin-S  & 93.60 & 96.18 & 85.75 & 90.36 & 87.64 & 90.71 & 91.63 & 83.22 \\
			HBCNet \cite{xu2022high}				& 	HRNet\_w48  & 93.60 & 96.13 & 85.95 & \textbf{90.53} & 90.40 & 91.32 & 91.72 & 84.21 \\
			\midrule
			
			TANet (Ours)  				& 	VGG-19  & 93.21 & 96.23 & 85.01 & 88.09 & 90.04 & 90.52 & 90.88 & 82.91 \\
			TANet (Ours)  				& 	ResNet-50  & 93.79 & 96.64 & 85.77 & 88.21 & 90.84 & 91.05 & 91.37 & 83.81\\
			
			TANet (Ours)  			& 	ResNet-101  & \textbf{94.16} & \textbf{96.80} & \textbf{86.95} & 88.84 & 90.52 & \textbf{91.45} & \textbf{91.93} & \textbf{84.45} \\
			
			\bottomrule
	\end{tabular}}
\end{table*}

\begin{table*}[htbp]
	\caption{\centering \textsc{Quantitative Comparisons with State of the Arts on the Potsdam Test Set}}
	\scalebox{1.05}{
		\begin{tabular}{l|c|c|cccc|c|c|c}
			\toprule
			& Backbone & Imp.surf &Building & Low veg & Tree  & Car & Mean F1(\%) & OA(\%) & mIoU(\%) \\
			\midrule
			UZ\_1 \cite{volpi2016dense} 			& -            & 89.30 & 95.40 & 81.80 & 80.50 & 86.50 & 86.70 & 85.80 & -     \\ 
			DANet \cite{fu2019dual}					& ResNet-101   & 91.96 & 96.35 & 86.20 & 87.21 & 95.92 & 91.48 & 89.98 & 84.57 \\
			V-FuseNet \cite{audebert2018beyond} 	& -            & 92.70 & 96.30 & 87.30 & 88.50 & 95.40 & 92.04 & 90.60 & -     \\ 
			TSMTA \cite{ding2020semantic}		 	& ResNet - 101 & 92.91 & 97.13 & 87.03 & 87.26 & 95.16 & 91.90 & 90.64 & -     \\
			TreeUNet \cite{yue2019treeunet} 		& -            & 93.10 & 97.30 & 86.60 & 87.10 & 95.80 & 91.98 & 90.70 & -     \\ 
			DeepLabV3+ \cite{chen2017rethinking} 	& ResNet - 101 & 92.95 & 95.88 & 87.62 & 88.15 & 96.02 & 92.12 & 90.88 & 84.32 \\ 
			CASIA3 \cite{liu2018semantic} 			& ResNet - 101 & 93.40 & 86.80 & 87.60 & 88.30 & 96.10 & 92.44 & 91.00 & -     \\
			PSPNet \cite{zhao2017pyramid} 			& ResNet - 101 & 93.36 & 96.97 & 87.75 & 88.50 & 95.42 & 92.40 & 91.08 & 84.88 \\ 
			MANet \cite{li2021multiattention} 		& ResNet - 50  & 93.40 & 96.96 & 88.32 & 89.36 & 96.48 & 92.90 & 91.32 & 86.95 \\ 
			CCANet \cite{deng2021ccanet} 			& ResNet - 101 & 93.58 & 96.77 & 86.87 & 88.59 & 96.24 & 92.41 & 91.47 & 85.65 \\
			HUSTW4 \cite{sun2019problems} 			& -            & 93.60 & 97.60 & 88.50 & 88.80 & 94.60 & 92.62 & 91.60 & -     \\ 
			MFNet \cite{su2022semantic} 			& ResNet - 50  & 94.25 & 97.52 & 88.42 & 89.43 & 96.62 & 93.25 & 91.96 & 87.57 \\ 
			HMANet \cite{niu2021hybrid} 			& ResNet - 101 & 93.85 & 97.56 & 88.65 & 89.12 & 96.84 & 93.20 & 92.21 & 87.28 \\  
			CTMFNet \cite{song2022ctmfnet}		& 	HRNet\&transformer  & 93.22 & 97.12 & 87.87 & 89.36 & 96.61 & 92.84 & 91.38 & 86.85 \\
			HBCNet \cite{xu2022high}				& HRNet\_w48   & 94.29 & 97.54 & 88.49 & 89.58 & 97.00 & 93.38 & 91.97 & 87.81 \\
			DC-Swin \cite{wang2022novel}		& 	Swin-S  & 94.19 & 97.57 & 88.57 & \textbf{89.62} & 96.31 & 93.25 & 92.00 & 87.56 \\
			\midrule
			
			TANet (Ours)  							& 	VGG-19  & 93.99 & 97.42 & 88.95 & 87.68 & 97.05 & 93.02 & 91.20 & 87.22 \\
			TANet (Ours)  							& 	ResNet-50  & 94.50 & 97.57 & 89.61 & 88.72 & 97.50 & 93.58 & 91.93 & 88.17\\
			
			TANet (Ours)  			& ResNet-101  & \textbf{94.65} & \textbf{97.65} & \textbf{89.80} & 88.97 & \textbf{97.54} & \textbf{93.72} & \textbf{92.45} & \textbf{88.41}  \\
		
			\bottomrule
	\end{tabular}}
\end{table*}

\textit{2) Efficiency Comparison}

We compare the efficiency of our proposed AFEM and TAPP modules with other contextual aggregation and attention modules in terms of parameters, GPU memory, and computational costs (GFLOPs). To ensure a fair comparison, as in \cite{li2021multiattention} and \cite{niu2021hybrid}, we use $3 \times 3$ convolutions for dimensionality reduction and evaluate the cost without considering the backbone cost. Table VIII displays the experimental results. Compared to the standard SA mechanism, the proposed module exhibits approximately 1/12 the GFLOPs, 1/2 the number of model parameters, and 1/8 the GPU memory. The optimized GFLOPs and number of model parameters demonstrate the superiority of our AFEM and TAPP modules compared to state-of-the-art methods.

Table IX presents the time cost of the model in the inference phase. In this experiment, the backbone of all models is set to ResNet-101. While our method may not have the most optimal time-cost performance, the time required by our model is not significantly different from other models. For instance, TANet required only 5 ms more than EANet and 3 ms more than DeeplabV3+. We consider the slight increase in time spent to be a reasonable tradeoff for obtaining improved segmentation results and significantly reducing the number of model parameters.

\captionsetup[figure]{labelformat=simple, labelsep=period}
\begin{figure}[htbp]
	\centering
	\includegraphics[scale=0.31]{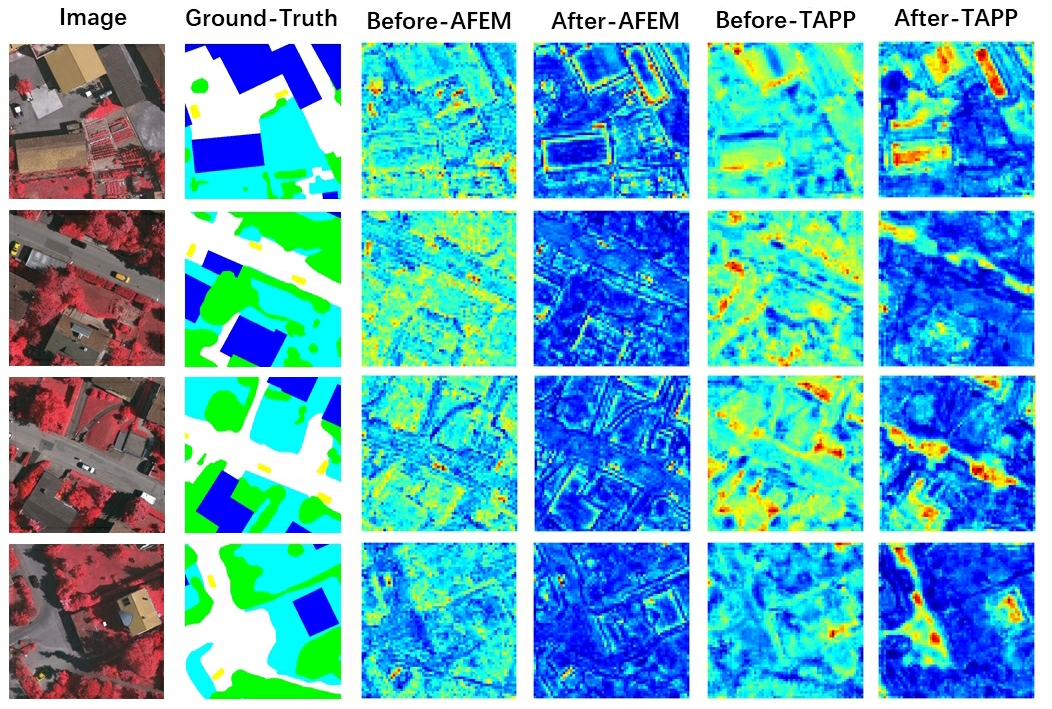}
	\caption{Visualisation of the AFEM module and TAPP module input and output features plotted on the Vaihingen test set.}
\end{figure}

\subsection{Comparison With State of the Art}
\textit{1) Experimental results on the Vaihingen dataset}

Similar to HMANet \cite{niu2021hybrid} and HBCNet \cite{xu2022high}, our proposed methods aim to enhance the model's segmentation results. Two improvements were made to the Loss function to enhance TANet network segmentation. The first is the Aux Loss, which accelerates convergence and improves segmentation outcomes. The second is the hard example mining (OHEM) method, which focuses the network more on challenging-to-classify pixels. We also employed the technique of Test-Time Enhancement (TTA). TTA involves flipping the input images horizontally and vertically during testing, leading to improved segmentation performance of the model. Our findings are presented in Table X and demonstrate that all three methods effectively improve the model's segmentation ability.

Table XII shows a comparison of our best segmentation results on the Vaihingen test set with state-of-the-art methods, including contextual aggregation methods and various attention-based methods. Our TANet uses ResNet-101 as the backbone, like most models. The results reveal that TANet outperforms the other methods, achieving the best results in all three important composite metrics. The experimental result supports the efficacy of our threshold attention mechanism and TANet architecture. We also experimented with adding AFEM and TAPP modules to the ResNet50 and VGG-19 backbones. The experimental results in Tables XII and XIII show that the addition of AFEM and TAPP modules on other different backbones can also effectively improve the segmentation of the model. The "-" symbol in the tables throughout this paper signifies the absence of data provided by the authors of the original paper. Additionally, reproducing their network model is challenging as the underlying code is not available as open source.

\textit{2) Visualisation of the attention module}

To enhance comprehension of the roles of AFEM and TAPP modules, which were designed based on TAM, the feature maps before and after these modules were visualized. The results are presented in Fig. 6. The AFEM module enhances edge differences between pixel blocks belonging to different objects, making object contours clearer and preserving more detailed information.

Comparing the before-TAPP and after-TAPP columns, it can be seen that the TAPP module enhances response value differences between regions belonging to different objects, making it easier for the model to distinguish semantic information. For instance, response values are relatively larger for both buildings and cars.

\captionsetup[figure]{labelformat=simple, labelsep=period}
\begin{figure*}[htbp]
	\centering
	\includegraphics[scale=0.52]{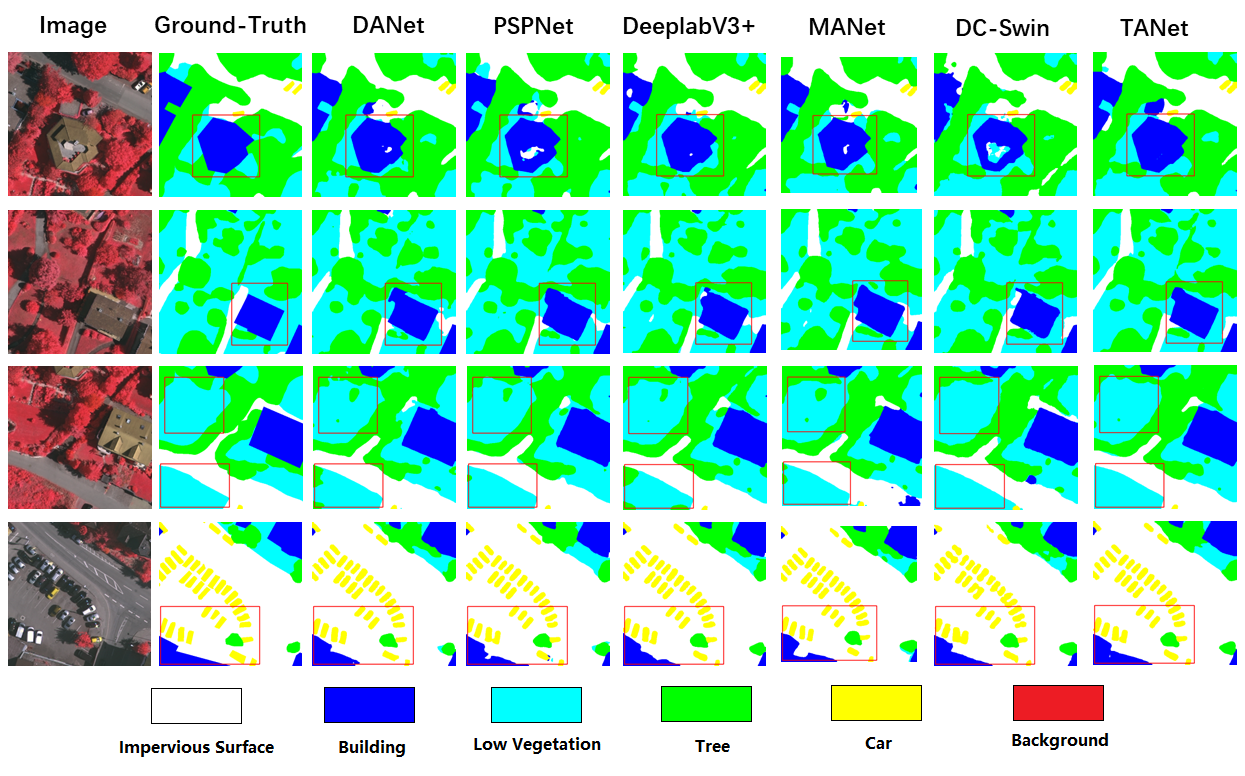}
	\caption{Qualitative comparison between our method (TANet) and other methods. The region in the red box represents a challenging segmentation area.}
\end{figure*}

\captionsetup[figure]{labelformat=simple, labelsep=period}
\begin{figure*}[htbp]
	\centering
	\includegraphics[scale=0.52]{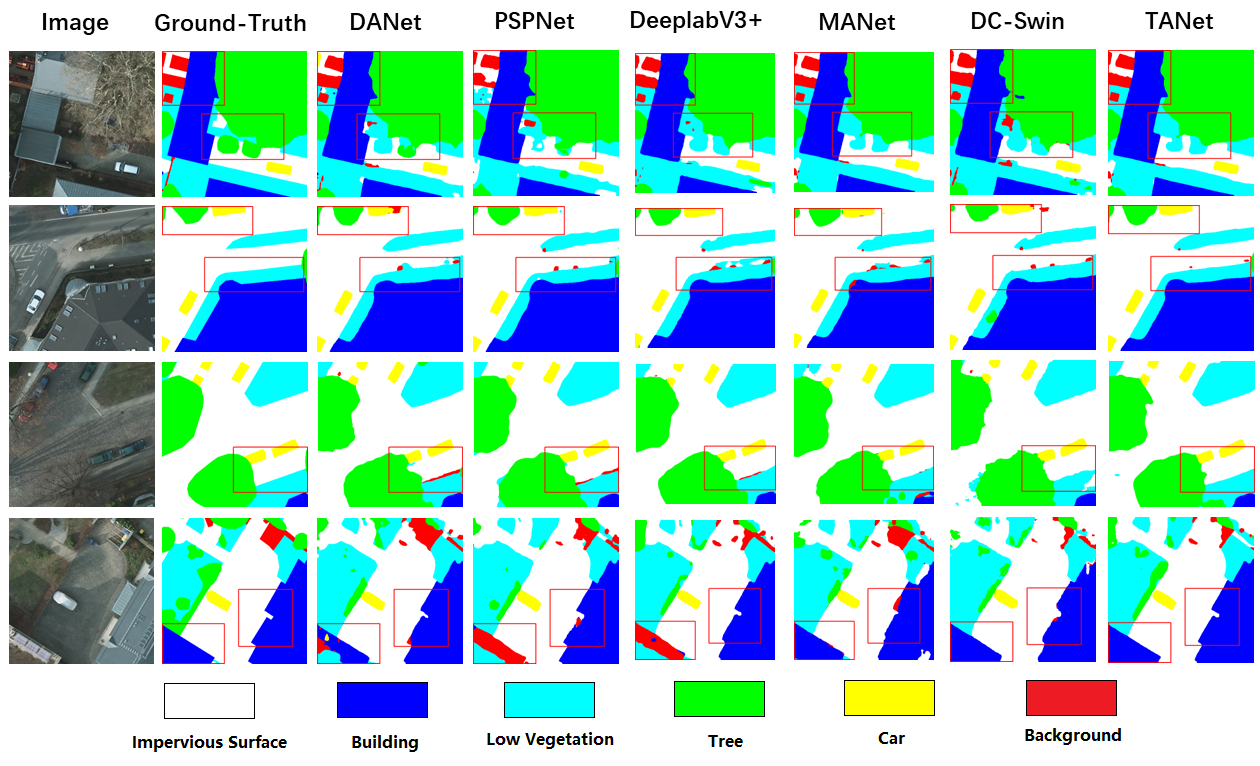}
	\caption{Qualitative comparison between our method (TANet) and other methods. The region in the red box represents a challenging segmentation area.}
\end{figure*}

\textit{3) Visualisation of results}

As shown in Fig. 7, we visualize the segmentation results of TANet on the Vaihingen test set and compare them qualitatively with several classical semantic segmentation networks. The region in the red box represents a challenging segmentation area. A comparison of the models' results clearly shows that TANet's predictions are the most similar to the true labeled maps in terms of object consistency and boundary definition. This emphasizes the effectiveness of the TAM in modeling pixel region features and enhancing object boundary details.

\textit{4) Experimental results on the Potsdam dataset}

To further assess the efficacy of TANet, experiments were conducted on the Potsdam dataset using the same three methods listed in Table XI as those performed on the Vaihingen dataset. Results were compared with the latest available methods and are shown in Table XIII. TANet achieved the highest scores in the three metrics of average F1, OA, and mIoU, outperforming all other models. Our method outperformed other approaches in most categories, with the exception of the TREE category. Further analysis suggests that this may be due to the thin branches and wide color distribution of the TREE category in the two datasets. These characteristics may make it difficult for our threshold attention method to accurately detect the region associated with this category. Fig. 8 visualizes TANet's segmentation results on the Potsdam test set, with the closest prediction to ground truth indicated in the red-boxed area.

\section{Conclusion}
In this paper, we propose a novel Threshold Attention Mechanism. In comparison to self-attention mechanisms, TAM significantly reduces computational effort while augmenting the correlation modeling between similar pixel regions in the feature map. Based on TAM, we design TANet, a semantic segmentation network for remote sensing images. TANet employs a pre-trained ResNet-101 as the backbone and extracts global relevance feature information from the deep network using the TAPP module. The shallow network output is augmented with region-specific feature information via the AFEM module, and the complementary information from both is subsequently combined to obtain the final prediction map. To validate our approach, we conducted experiments on two high-resolution remote sensing image semantic segmentation datasets, Vaihingen and Potsdam. The results show that TANet outperforms other methods in most overall metrics, demonstrating the efficacy of our approach.

\bibliographystyle{IEEEtran}
\bibliography{IEEEabrv,Threshold_Attention_Network_for_Semantic_Segmentation_of_Remote_Sensing_Images}

\vspace{-35pt}

\end{document}